\documentclass[10pt,twocolumn,letterpaper]{article}

 \usepackage[pagenumbers]{cvpr} 

\definecolor{cvprblue}{rgb}{0.21,0.49,0.74}
\usepackage[pagebackref,breaklinks,colorlinks,allcolors=cvprblue]{hyperref}
\usepackage{graphicx}

\DeclareGraphicsExtensions{.eps}
\usepackage{array}
\usepackage{bbding}
\usepackage{pifont}
\usepackage{bm}
\usepackage{cases}
\usepackage{amssymb}

\usepackage{amsthm}
\usepackage{hyperref}

\usepackage{tabulary}%
\usepackage{tabularx}
\usepackage{bm}
\usepackage{url}  
\usepackage{graphicx}  
\usepackage{amsmath}
\usepackage{array}
\usepackage{amsfonts}
\usepackage{tabularx}
\usepackage{epstopdf}
\usepackage{algorithm} 
\usepackage{algorithmic}
\usepackage{multirow} 
\usepackage{booktabs}
\usepackage{color}

\usepackage{flushend} 
\usepackage{ulem}    

\newfloat{listing}{tb}{lst}{}
\floatname{listing}{Listing}
\usepackage{CJKutf8}
\def\paperID{10680} 
\def\confName{CVPR}
\def\confYear{2025}

\title{Relation-Aware Meta-Learning for Zero-shot Sketch-Based Image Retrieval}

\author{
Yang~Liu$^{1}$, Jiale~Du$^{1}$, Xinbo~Gao$^{1,2}$, Jungong~Han$^{3}$ \\
$^{1}$Xidian University, Xi'an, China \\
$^{2}$Chongqing University of Posts and Telecommunications, Chongqing, China \\
$^{3}$Tsinghua University, Beijing, China \\
{\tt\small yangl@xidian.edu.cn, 23011211070@stu.xidian.edu.cn, xbgao@mail.xidian.edu.cn, }\\{\tt\small jungonghan77@gmail.com}
}

\begin{document}
\begin{CJK}{UTF8}{gbsn}
\maketitle
\begin{abstract}
Sketch-based image retrieval (SBIR) relies on free-hand sketches to retrieve natural photos within the same class. However, its practical application is limited by its inability to retrieve classes absent from the training set. To address this limitation, the task has evolved into Zero-Shot Sketch-Based Image Retrieval (ZS-SBIR), where model performance is evaluated on unseen categories. Traditional SBIR primarily focuses on narrowing the domain gap between photo and sketch modalities. However, in the zero-shot setting, the model not only needs to address this cross-modal discrepancy but also requires a strong generalization capability to transfer knowledge to unseen categories. To this end, we propose a novel framework for ZS-SBIR that employs a pair-based relation-aware quadruplet loss to bridge feature gaps. By incorporating two negative samples from different modalities, the approach prevents positive features from becoming disproportionately distant from one modality while remaining close to another, thus enhancing inter-class separability. We also propose a Relation-Aware Meta-Learning Network (RAMLN) to obtain the margin, a hyper-parameter of cross-modal quadruplet loss, to improve the generalization ability of the model. RAMLN leverages external memory to store feature information, which it utilizes to assign optimal margin values. Experimental results obtained on the extended Sketchy and TU-Berlin datasets show a sharp improvement over existing state-of-the-art methods in ZS-SBIR.
\end{abstract}
\section{Introduction}
\label{sec:intro}

\begin{figure}[!h]
\centering
\includegraphics[width=3.3in,height=2.3in]{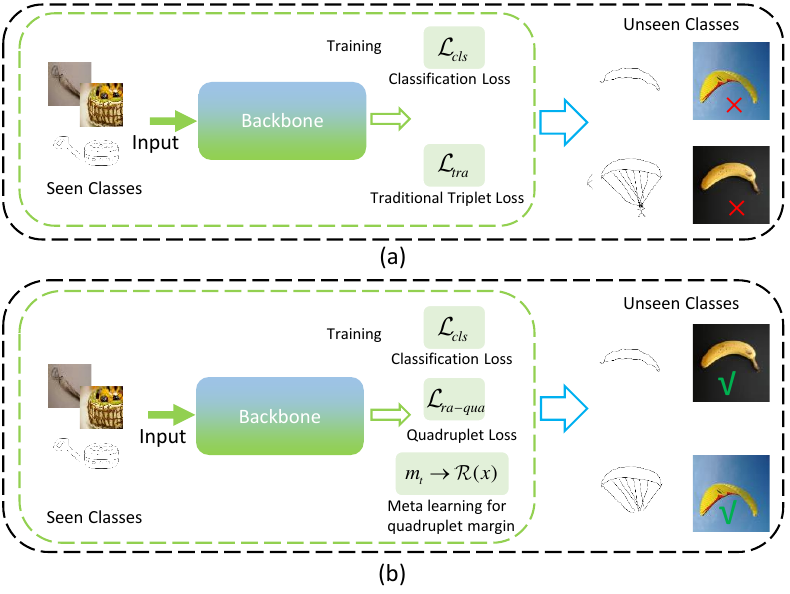}
\caption{(a) and (b) respectively illustrate the performance of the traditional triplet loss method and our proposed approach in handling the ZS-SBIR task. The traditional triplet loss method tends to misclassify objects with similar shapes but belonging to different categories. For example, due to their similar shapes, it might mistakenly classify a parachute as a banana. In response, we propose a novel relation-aware quadruplet loss function to thoroughly explore both cross-modal and intra-modal relationships. Additionally, we employ a meta-learning strategy to optimize the margin in quadruplet loss, adaptively determining the optimal margin value. This approach not only enhances the model's generalization ability but also significantly improves its capacity to distinguish between objects with similar shapes, such as accurately differentiating a parachute from a banana.}

\label{intro}
\end{figure}
Sketch-based image retrieval (SBIR) \cite{sangkloy2016sketchy} aims to retrieve photos based on the queries of sketches. It is of significant value on touch-screen devices. However, it is very difficult to have all categories of the training set cover all query categories at the application stage. Shen \emph{et al.} \cite{shen2018zero} combine SBIR with zero shot setting, and propose zero shot sketch-based image retrieval (ZS-SBIR). ZS-SBIR requires retrieving photos with the query sketches whose categories have not appeared in the training set, \emph{i.e.}, training and testing set have no class intersection. Therefore, it has more convenient application scenarios. Similar to SBIR tasks, many studies treat ZS-SBIR as a metric learning, where sketches and photos are mapped to a shared latent embedding space, and their similarity is measured by calculating the distance between their features.

The ZS-SBIR task faces the same fundamental challenge as SBIR, namely, the substantial modal gap between sketches and photos. Bridging this gap is challenging, as it complicates the model’s ability to capture shared visual details across modalities. Consequently, samples within a single class tend to separate into two distinct clusters. Unlike conventional SBIR, however, ZS-SBIR demands not only a cohesive grouping of intra-class samples but also a reduction in the inter-group distance between modalities. Based on this, previous works have designed different frameworks, such as GANs~\cite{dutta2019semantically}, graph~\cite{zhang2020zero}, and cycle reconstruction~\cite{han2019learning}. These works learn the projections of photos and sketches in the embedding space, which has advantages in the measurement of distance. In addition, ZS-SBIR suffers from the domain gap between seen and unseen classes. The distribution of seen classes is different from that of unseen classes. The model fails to transfer the knowledge learned from the seen classes to the unseen classes and is unable to align the two domains, leading to the domain shift problem. To solve this problem, doodle2search \cite{dey2019doodle} uses triplet to reduce the distance between embedded sketch and photo if they belong to the same class and increase it if they belong to different classes. It only uses one triplet: sketch as the anchor while photo as positive/negative sample. SBTKNet \cite{tursun2022efficient} considers the inter-class and intra-class distance relationship. The two negative samples come from different modalities, and the negative pair in the same modality as the anchor is the main pair, which can further distance them. 

We aim to investigate the impact of utilizing photos as anchor points on the alignment between photos and sketches, as well as on the construction of the embedding space. Anchors for both modalities help to explore the inter-modal and intra-modal alignment relations. On the other hand, different triplets correspond to different distances and relationships. We hope to find the different margin, which is the hyperparameter of triplet loss. We believe that a strategy grounded in features and global information could potentially yield effective results. In detail, we propose a novel model named Relation-Aware Meta-Learning Network (RAMLN) that adopts relation-aware quadruplet loss to construct generalizable embedding space relationship. Four samples are combined in pairs to form three pairs for contrastive learning. When the anchor is far away from the primary negative sample, the secondary negative sample helps stabilize the anchor position. We also employ a distance-based hard mining strategy. Instead of centering on a class group, it directly pushes the closest samples and pulls the farthest samples, which is more sensitive to the distance. At last, we decide to learn the margin of quadruplet loss inside a meta-learning. It is key for ZS-SBIR to use the seen data to improve the generalization ability of the model in the unseen domain. An external memory space is used to assist unseen features by recording uncommon memory in seen samples. So we adopt meta-learning with a memory-augmented network to adjust the margin parameters in the quadruplets to capture important rare features.

The main contributions of this work are summarized as follows:
\begin{itemize}
  \item We propose a novel relation-aware quadruplet loss to mine the inter-modal and intra-modal relation. Two negative samples from different modalities ensure that the anchor can avoid both non-similar seen and unseen domains.
  \item We propose a meta-learning approach to learn the margin in quadruplet loss, adaptively determining the optimal margin value. The adaptive margin not only alleviates issues with improper margin settings but also accommodates domain variations across different categories and modalities.
  \item We demonstrate the validity and high performance of the proposed model by conducting experimental evaluations on two popular ZS-SBIR datasets Sketchy and TU-Berlin.
\end{itemize}
\section{Related Work}
\label{sec:Related}
\subsection{Zero-Shot Sketch-Based Image Retrieval}

Zero-shot learning is a more challenging and practically significant task, requiring the model to handle samples from classes that were not present in the training set. It is a subtask of transfer learning, aiming to effectively transfer knowledge from the seen domain to the unseen domain. ZS-SBIR performs SBIR in the zero-shot setting: the test class of ZS-SBIR does not appear in the training phase. Shen \emph{et al.} \cite{shen2018zero} first introduced the SBIR problem under the zero-shot setting. Similar to SBIR, ZS-SBIR is a challenging task that addresses the modality gap between sketches and natural photos. Most works try to map the sketches and photos into an embedding space, and obtain the suitable position and distance of sample features in the space through metric learning, such as IIAE \cite{hwang2020variational}, SBTKNet \cite{tursun2022efficient} and so on. Jing \emph{et al.} \cite{jing2022augmented} propose a novel Augmented Multimodality Fusion (AMF) framework that employs a knowledge discovery module to mimic novel knowledge unseen during the training phase, which helps train the model to adapt to the gap from the seen domain to the unseen domain. Moreover, many works have also tried to map semantic information to the embedding space as side information. It provides additional semantic knowledge to bridge the gap and localize the features of unseen classes in space. The earlier works like doodle2search \cite{dey2019doodle, hu2013performance} and recent works like ocean \cite{zhu2020ocean}, SkechGCN \cite{zhang2020zero, dutta2020adaptive, song2017deep}  adopt with language models, such as Word2vec, Bag-of-words and text transformer, to obtain text vectors through semantic information. But this does not mean that semantic information is necessary. Wang \emph{ et al.} \cite{wang2021transferable} propose a novel Transferable Coupled Network (TCN) to improve the transferability. To better utilize semantic information, TCN elaborates a semantic metric to integrate local metric learning and global semantic constraints into a unified formulation. Lin \emph{et al.} \cite{lin2023zero} proposed a patch matching framework ZSE by a cross-attention module to compute local correspondences between tokens across two modalities. Without external semantic knowledge, ZSE still outperforms the aforementioned methods. ABDG \cite{tian2023zero} uses multiple teacher models for knowledge distillation. There are two teacher models for sketches and photos to improve the model's generalization ability, and one teacher model for ZS-SBIR task to improve the discrimination ability. Then it balances these different distillations using an entropy-based approach. ABDG has achieved very good performance, but it costs a lot.

\begin{figure*}[!t]
\centering
\includegraphics[width=6.9in,height=3.65in]{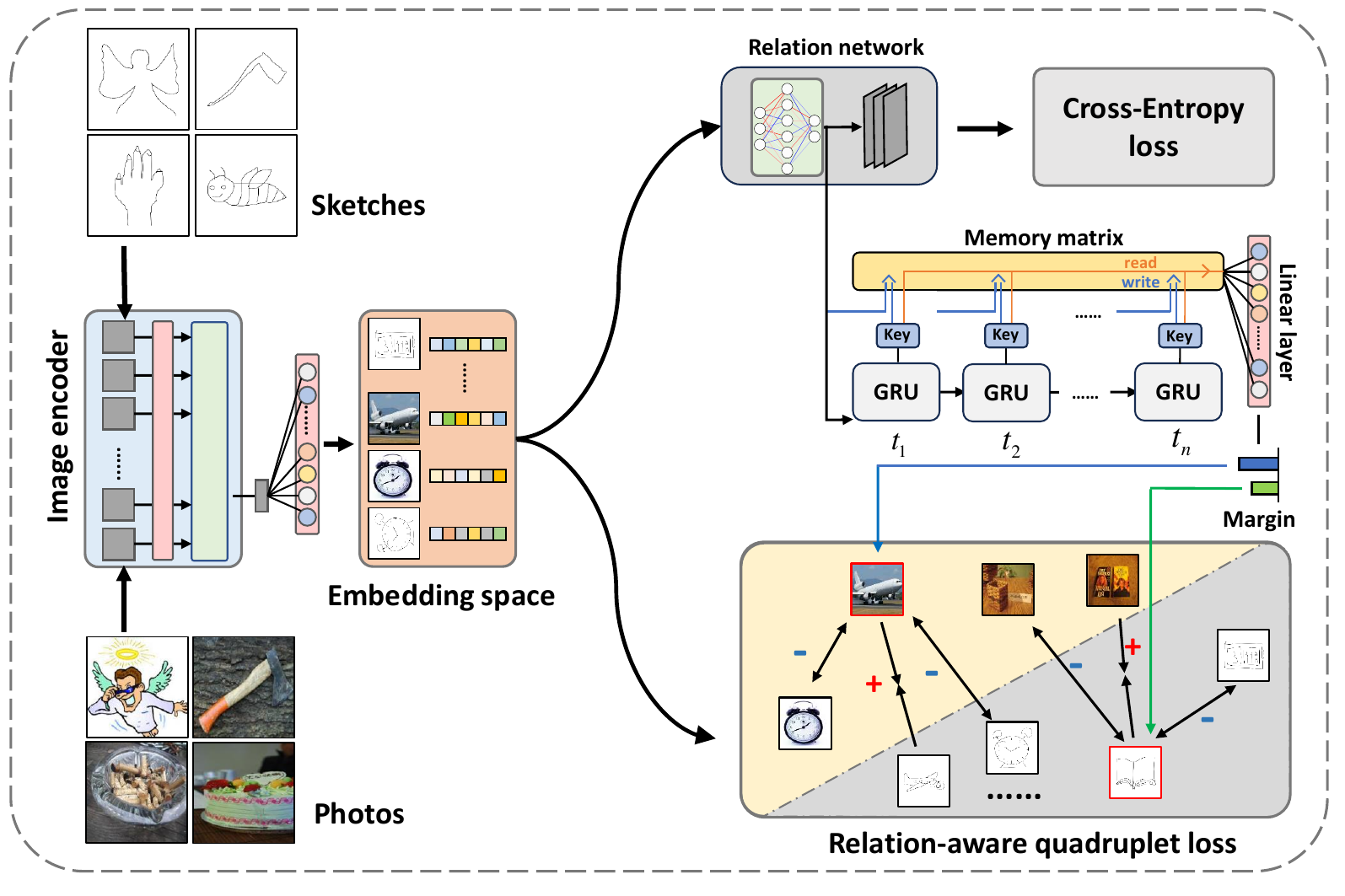}
\caption{The overall structure of the proposed method. The image encoder extracts features from both sketches and photos in the embedding space. Then the training architecture combines two parts: (i) Bidirectional training, which incorporates relation-aware quadruplets from different modalities to learn feature distributions, and optimizes using the margin obtained through meta-optimization. (ii) Classification training utilizes cross-entropy loss with the Softmax function, which helps to avoid getting stuck in local optima. Retrieval is based only on distance, classification training still makes sense.}
\label{fig2}
\end{figure*}

\subsection{Meta-Learning}
Meta-learning \cite{9665225}, often described as “learning to learn,” enables models to acquire new tasks by leveraging prior knowledge. It has emerged as a key direction for advancing AI performance and tackling complex challenges. The approaches to meta-learning are highly diverse, encompassing a range of distinct methodologies. For instance, Marcin \emph{et al.} \cite{andrychowicz2016learning} propose training a neural network to predict gradients. Similarly, Sung \emph{et al.} \cite{sung2017learning} introduce a Meta-Critic network to predict the main network's loss, analogous to the q-value in reinforcement learning. Vinyals \emph{et al.} \cite{vinyals2016matching} develop an attention model that directly focuses on essential components in new tasks. The memory-based method \cite{santoro2016meta}, a frequently used approach, employs an external memory to assist the network in retaining and adapting to new classes with minimal samples. This technique is widely applied in anomaly detection \cite{park2020learning}, object tracking \cite{yang2018learning}, and person re-identification \cite{zhong2020learning}. Notably, however, meta-learning has yet to be applied to the ZS-SBIR task. To enable the model to generate new margin values, it is logical to look beyond current data and incorporate insights from historical experience. Our aim is to construct a robust memory that is dynamically read and written based on current data, where global information guides more informed decision-making.
\section{Proposed Approach}
The dataset of ZS-SBIR can be divided into seen classes and unseen classes. Seen classes are used as the train set while unseen classes are used as the validation set and test set. The train set is denoted by $ X_{seen}=\left\{\left(x_{i}^{m}, c_{i}\right) \mid c_{i} \in C_{seen}, m \in\{ske, pho\}\right\}_{i=1}^{N} $, where $x$ is feature extracted from images and $c$  is label respectively. $m$ is the modality with sketchy $ske$ and photo $pho$. $C_{seen}$ denotes the seen classes set. $N$ is the number of seen samples. Similarly, the test set is denoted by $ X_{unseen}=\left\{\left(x_{j}^{m}, c_{j}\right) \mid c_{j} \in C_{unseen}, m \in\{ske, pho\}\right\}_{j=1}^{M} $.  To satisfy the zero-shot setting, $C_{seen}\cap C_{unseen}=\emptyset $.

As shown in Figure. \ref{fig2}, key components of the proposed model RAMLN include: (i) With an additional linear layer at the end of the image encoder, the features of photos and sketches are mixed as its input. The features within the embedding space are meticulously prepared to facilitate downstream tasks. (ii) Memory-based meta-optimization for the margin of quadruplet, which is a hyper-parameter. Through an additional memory matrix, the model can adapt to the unseen domain. (iii) Selected relation-aware quadruplet loss based on the normalized Euclidean distance. (iv) A classification function connected with a projection layer, where the output dimension is the number of the class.

\subsection{Triplet Loss} 
Before presenting our approach, we need to first introduce the notion of triplet loss and analyze it. Triplet loss uses three samples to form two pairs: anchor samples, positive samples of the same class, and negative samples of different classes. Anchor samples and positive samples form positive pairs, and anchor samples and negative samples form negative pairs. By optimizing the intra-class and inter-class distances through positive and negative pairs, the model can learn a good feature embedding space. The design of triplet loss is straightforward yet highly effective, making it widely applicable across various metric learning tasks. The target order for conventional triplet loss is:
\begin{equation}\label{tri}
\mathcal{L}_{tri}=D\left(x_{a}^{}, x_{p}^{}\right)-D\left(x_{a}^{}, x_{n}^{}\right)+\mu,
\end{equation}
where $\mu$ is a given margin for better separability in the embedding space for different categories and $D$ is the distance between two image features. $x_{a},x_{p},x_{n}$ denote anchor, positive sample and negative sample, respectively.

When constructing triplets to meet task-specific requirements, the sketch generally serves as the anchor, while photos are chosen as the positive and negative samples. For ZS-SBIR tasks that involve two distinct modalities, however, this conventional triplet structure proves overly simplistic. It fails to consider the optimization of other distance metrics, potentially leading to an unbalanced feature distribution within the embedding space. Additionally, the sample distribution between different modalities is not symmetric, and random selection of positive pairs $ D\left(x_{a}^{}, x_{p}^{}\right)$ and negative pairs $D\left(x_{a}^{}, x_{n}^{}\right)$ may result in suboptimal distributions. To address these issues, we propose a relation-aware quadruplet loss and strategically select pairs tailored for the ZS-SBIR task.

\subsection{The relation-aware quadruplet loss}
Considering the relation-aware nature of the SBIR task, we introduce the triplet loss into two domains and improve the triplet loss to a quadruplet loss. With pairs across different modalities, the quadruplet loss pushes the samples to the appropriate position in the embedding space. We use the Euclidean distance $D\left(.\right)$  between two images to measure their dissimilarity. It is unreasonable to only measure certain combinations like Eq. \ref{tri}. To address the absence of a unified distance metric for negative pairs, our relation-aware quadruplet loss incorporates a second negative pair as an additional constraint. The quadruplet loss can be formulated as follows:
\begin{equation}\label{eq2_1}
\begin{split}
\mathcal{L}_{qua}=&D\left(x_{a}^{}, x_{p}^{}\right)-\left(1-\lambda\right)D\left(x_{a}^{}, x_{n_{1}}^{}\right)\\
&-\lambda D\left(x_{a}^{}, x_{n_{2}}^{}\right)+\mu,\\
\end{split}
\end{equation}
where $x_{a}$, $x_{p}$, $x_{n_{1}}$ and $x_{n_{2}}$ represent anchor image, positive image, the first negative image and the second negative image.

In contrast to the triplet loss, the new second negative term is the distance between the anchor and the second negative image. Similar to the triplet loss, the first term obtains non-identical class distances and acts as a ``strong push'' to increase the inter-class gap. However, a single negative gradient direction does not necessarily do this. Thus the second term assists in providing another gradient direction for updating. To make the quadratic loss comprehensive, we intuitively set the second negative image from a different modality than the negative image in the first negative term. This auxiliary term helps ``strong push'' to properly increase the inter-class gap, so it plays the role of ``weak push'' and we use weight $\lambda$ to ensure the primary and secondary relation between the two constraints. Since we use Euclidean distance squared as the metric, the back-propagation of a triplet is:
\begin{equation}\label{back}
\begin{split}
-\frac{\partial Tri}{\partial x_{a}}=-2(x_{n}-x_{p}), \\
-\frac{\partial Tri}{\partial x_{p}}=-2(x_{p}-x_{a}), \\
-\frac{\partial Tri}{\partial x_{n}}=-2(x_{n}-x_{p}). \\
\end{split}
\end{equation}

However, following Eq. \ref{back}, the negative gradient direction may not be optimal. So it may be impossible to separate positive and negative sample centers. 

We strategically designed the selection of pairs to effectively enhance inter-class distances and diminish intra-class distances, particularly by bridging the gap between different modalities. First, considering optimizing the whole embedding space instead of optimizing the position of sketches relative to photos, we carry out experiments for the selection of various pairs and extend anchors to photos. Second, since the gap between photos and sketches is larger than the distance between images in the same modality. So we do not need to concentrate on increasing the distance across modalities, but reducing the distance of the same class, which means that for samples of the same class, ``pull'' within the class and ``push'' within the same modality are more effective than ``push'' between modalities. Last, we adopt the hard mining strategy, utilizing only hard positive pairs (the closest negative sample and anchor) and hard negative pairs (the farthest positive sample and anchor) to train the model. This helps to filter out useless easy pairs and speed up the convergence of the model.

Based on the above reasons, we obtain a uniform setting for negative pairs. Indeed, dividing by the positive sample, we have two quadruplets in our relation-aware quadruplet: a global inter-modal quadruplet and a local intra-modal quadruplet. In ZS-SBIR, samples from the same class in the embedding space are always split into two parts due to natural differences between sketches and photos. Commonly, two parts are further apart than one part itself. So the main task in ZS-SBIR is to reduce the gap between sketch and photo. Our global inter-modal quadruplet focuses on this task, using a relation-aware positive pair to reduce the relation-aware distance. For simplify, we use $\mathcal{P}_{\alpha,\beta}^{+}(x)$ and $\mathcal{P}_{\alpha,\beta}^{-}(x)$ to represent positive and negative pairs:
\begin{equation}\label{pair}
\begin{split}
 \left\{\begin{array}{ll}
\mathcal{P}_{\alpha,\beta}^{+}(x)= max(D(x_{a}^{m_{\alpha}},x_{p}^{m_{\beta}})) \\
\mathcal{P}_{\alpha,\beta}^{-}(x)= min(D(x_{a}^{m_{\alpha}},x_{n}^{m_{\beta}})).
\end{array}\right.
\end{split}
\end{equation}

We use $m_{\alpha}$ and $m_{\beta}$ to represent the modality of the anchor $x_{a}$ , positive sample $x_{p}$ and negative sample $x_{n}$. Then the global inter-modal quadruplet is:
\begin{equation}\label{inter}
\begin{split}
&\mathcal{L}_{inter}(x,j,k)= \mathcal{P}_{j,k}^{+}(x)+\mathcal{R}(x)\\
&\ \ \ \ \ \ \ \ \ \ \ \ \ \ \  \ \ \ \ \ -[\lambda \mathcal{P}_{j,k}^{-}(x)+(1-\lambda) \mathcal{P}_{j,j}^{-}(x)] ,
\end{split}
\end{equation}
where $j\neq k$. Since we both adopt sketches and photos as the anchor, modalities $m_{j}$ and $m_{k}$ are not a one-to-one correspondence between sketches and photos. $\lambda$ is a balancing weight. It controls the ``strong push'' and ``weak push''. $\mathcal{R}\left(.\right)$ is the margin for different quadruplet obtained by meta-learning.

For samples in the same modality, the distribution is still scattered and there are many outlier samples at the beginning of the training. Therefore, we use the local intra-modal quadruplet to enhance the compactness of two parts separately. Its positive images are in the same modality of the anchor. While pushing away the negative group, it helps two groups of the same class reduce their respective inter-class distances:
\begin{equation}\label{qua1}
\begin{split}
&\mathcal{L}_{intra}(x,j,k)= \mathcal{P}_{j,j}^{+}(x)+\mathcal{R}(x)\\
&\ \ \ \ \ \ \ \ \ \ \ \ \ \ \  \ \ \ \ \ -[\lambda \mathcal{P}_{j,k}^{-}(x)+(1-\lambda) \mathcal{P}_{j,j}^{-}(x)].
\end{split}
\end{equation}

Different from loss (\ref{qua1}), we have the same $m_{j}$ of $x_{a}$ and $x_{p}$. By changing the anchor from the sketch and the photo, the intra-modal quadruplet works for the corresponding modality. It makes up for the inability of inter-modal loss (\ref{qua1}) to shorten the distance of samples. Two quadruplets form relation-aware quadruplet and comprehensively adjust the position of samples in the embedding space. Based on the above, our relation-aware quadruplet loss can be expressed as follows:
\begin{equation}\label{quar}
\begin{split}
\mathcal{L}_{ra-qua}=\sum_{j=1}^{2}\sum_{k=1,k\neq j }^{2} \mathcal{L}_{inter}(x,j,k)+\mathcal{L}_{intra}(x,j,k).
\end{split}
\end{equation}

\subsection{Meta Optimisation for the Loss Margin}
The traditional triplet loss contains the hyper-parameter, \emph{i.e.}, margin $\mu$ in Eq. \ref{tri}, whose optimal value is typically determined empirically and can vary across different categories. Given that the intra-class distribution or spread among sampled sketches may not be uniform across classes, it is intuitively reasonable to define a category-specific optimal margin. This applies equally to the boundary terms in our proposed quadruplet loss, namely $\mathcal{R}(x)$ in Eqs. \ref{inter} and \ref{qua1}. Therefore, we introduce a meta-learning process to learn this margin hyperparameter, allowing the optimal value of $\mathcal{R}(x)$ for each specific category to be adaptively determined during testing. Specifically, we employ a relation network with a controller to achieve this. This network inputs each row vector into each time step of a bidirectional Gated Recurrent Unit (GRU) to model the relationships among all samples in the training dataset. Subsequently, we apply max-pooling to the outputs from all time steps. The resulting vector is then fed into a linear layer, which ultimately outputs a sigmoid-normalized scalar value representing the learnable margin. The final memory vector $m_{t}$ is obtained by weighted summation:
\begin{equation}\label{weight}
\begin{split}
m_{t} = \sum_{i} w_{t}^{r}(i) M_{t}(i).
\end{split}
\end{equation}

The matrix being read $m_{t}$ is used for margin $\mathcal{R}(x)$. Considering $\mathcal{L}_{ra-qua}$ contains $\mathcal{L}_{inter}$ and $\mathcal{L}_{intra}$, our method learns margin $\mathcal{R}(x)$ which has two values. Thus, We use a $(dim,2)$ linear layer with ReLU to obtain the margin. Learning two different margin helps $\mathcal{L}_{inter}$ and $\mathcal{L}_{intra}$ learn different relationship. The margin derived through meta-optimization is deemed more suitable than a fixed margin. It combines the advantages of long-term memory and short-term memory to match the current features with the vectors in the external memory matrix and makes reasonable predictions of margin values. A detailed explanation of the margin optimization in meta-learning, along with the corresponding formula derivations, can be found in the supplementary materials.

\subsection{Classification Loss}
Our model computes the negative Euclidean distance between photos and sketches as the final similarity score. But we still use a linear layer to learn another similarity score from the output vectors and feed them into the classification loss to avoid getting trapped in bad local optimum. In our model, the standard cross-entropy loss combined with Softmax is used as the classification loss. The cross-entropy loss employs an inter-class competition mechanism and only cares about the accuracy of the predicted probability of the correct label. This helps the model to learn good inter-class relations and stabilizes the training process to avoid getting trapped in bad local optimum. In our model, the standard cross-entropy loss is used to determine how close the actual output is to the target, and the Softmax function provides the required probability for the cross-entropy loss in multi-class classification. From this, the classification loss can be formulated as follows:
\begin{equation}\label{cls}
\mathcal{L}_{cls }=- \sum_{i=1}^{N} c_{i}\log [\operatorname{softmax}(W x+b)],
\end{equation}
where $W$ and $b$ are the weight matrix and bias vector of the softmax cross-entropy loss, respectively. $c_{i}$ is the class label. The cross-entropy loss only focuses on whether the prediction is correct or not, ignoring correlations between classes or modalities. 

\begin{algorithm}[!h]
\caption{\textbf{: Training procedure of RAMLN}}
\label{alg:algorithm}
\textbf{Input}: Seen samples $ {\cal I}_{seen}=\left\{i_{i}^{m}, l_{i}\right\}  $, hyper-parameters $\lambda $, $\beta $, $\varphi $, batch size $B$ , learning rate $\mu $ and backbone  learning rate multiplier $
\alpha$.\\
\textbf{Output}: Networks parameters $\bf{\theta}$\\
\textbf{Training}
\begin{algorithmic}[1] 
\STATE Initialization networks parameters $\bf{\theta}$ and memory matrix.
\REPEAT
\STATE Randomly sample images in ${{\cal I}_{seen}}$ with batch.
\STATE Obtain features $x$ by backbone model.
\STATE Obtain key $k_{t}$ for memory matrix by GRU.
\STATE $m_{t} \leftarrow\sum_{i} w_{t}^{r}(i) M_{t}(i)$ Read vectors.
\STATE Fusion features and give margin $\mathcal{R}(x)$.
\STATE $M_{t}(i)\leftarrow M_{t-1}(i)+\textbf{w}_{t}^{w}(i)k_{t}$ Write vectors.
\STATE Solve $\mathcal{L}_{ra-qua }$ using Eq. (\ref{quar}).
\STATE  Solve $\mathcal{L}_{cls }$ using Eq. (\ref{cls}).
\STATE Update ${\theta }$ using ${\theta} \leftarrow {\theta } - \bigtriangledown_{\theta }(\mathcal{L}_{cls}+\lambda \mathcal{L}_{qua})  $.
\UNTIL{Max training epochs is reached.}\\
\end{algorithmic}
\end{algorithm}

\begin{table*}[!t]
\centering
\caption{The performance comparison of our method and competitors on Sketchy Extended dataset and TU-Berlin Extended dataset. $"-"$ denotes that results are not reported in the original papers. In ZS-SBIR, below the line are  ViT-based methods while above the line are the methods using Resnet and its variants as backbone or others. The best results are in \textcolor{red}{red}, and the second results are in \textcolor{blue}{blue}.}
\label{tab-marks}
\resizebox{\textwidth}{!}{
\begin{tabular}{*{9}{c}}
  \toprule
  \multirow{2}*{Methods} &\multirow{2}*{semantic} & \multirow{2}*{Dim} & \multicolumn{2}{c}{TU-Berlin Ext.}&\multicolumn{2}{c}{Sketchy-No Ext.}&\multicolumn{2}{c}{Sketchy Ext.}\\
  \cmidrule(lr){4-5}\cmidrule(lr){6-7}\cmidrule(lr){8-9}  
  &&    & mAP@all &  Prec@100  &mAP@200& Prec@200 &mAP@all& Prec@100 \\

  \midrule
    SAKE(ICCV-19)\cite{kodirov2017semantic}& \Checkmark&  512&  0.475&0.599&0.497 &0.598&0.547&0.692\\
    NAVE(IJCAI-21)\cite{wang2021norm}& \XSolidBrush &512& 0.493&0.607&-&-&\textcolor{blue}{0.613}&0.725\\

    Doodle(CVPR-19)\cite{dey2019doodle} &\Checkmark&4096&0.109&-&0.461&0.370&-&-\\

    SEM-PCYC(CVPR-19)\cite{dutta2019semantically}& \Checkmark& 64&0.297&0.426&-&-&0.349&0.463\\

     RPKD(ACM MM-21)\cite{tian2021relationship}& \XSolidBrush& 512&0.486&\textcolor{blue}{0.612}&0.502&0.598&\textcolor{blue}{0.613}&0.723\\

    Sketch3T(CVPR-22)\cite{sain2022sketch3t}& \XSolidBrush&512& \textcolor{blue}{0.507}&-&-& \textcolor{blue}{0.624}&-&-\\

    DSN(IJCAI-21)\cite{wang2021domain}& \Checkmark&512&0.481&0.586&-&-&0.583&0.704\\


         PSKD(ACM MM-22)\cite{wang2022prototype}& \XSolidBrush& 512&0.491&0.601&0.516&0.609&0.596&\textcolor{blue}{0.732}\\
         CA(TCSVT-23)\cite{wang2023cross}& \XSolidBrush& 512&0.482&0.594&-&-&0.590&0.713\\
        \textbf{RAMLN(ResNet)}&\XSolidBrush&512& \textcolor{red}{0.509}&\textcolor{red}{0.615}& \textcolor{blue}{0.524}& \textcolor{red}{0.630}&\textcolor{red}{0.630}&\textcolor{red}{0.741}\\
        \midrule
    TVT(AAAI-22)\cite{tian2022tvt}& \XSolidBrush&384& 0.484&  0.662& 0.531&0.618&0.648&0.796\\

    PSKD(ViT)(ACM MM-22)\cite{wang2022prototype}& \XSolidBrush& 512&0.502&0.662&{0.560}&{0.645}&0.688&0.786\\
    SASA(SIGIR-22)\cite{tian2022structure}& \XSolidBrush& 512&0.488&{0.670}&0.531&0.618&-&-\\

    ZSE-RN(CVPR-23)\cite{lin2023zero}&\XSolidBrush &512&0.542& 0.657&0.525&0.624&0.698&0.797 \\
    ZSE-Ret(CVPR-23)\cite{lin2023zero}&\XSolidBrush &512&\textcolor{blue}{0.569}& 0.637&0.504&0.602&{0.736}&{0.808} \\
    IVT(AEI-24)\cite{ZHANG2024102398}&\XSolidBrush&384& 0.557&\textcolor{blue}{0.692}&\textcolor{blue}{0.615}&\textcolor{blue}{0.694}&\textcolor{blue}{0.751}&\textcolor{red}{0.867}\\
    AMA(AAAI-24)\cite{yin2024asymmetric}&\XSolidBrush&-& 0.429&0.592&0.491&0.585&0.548&0.684\\
    CMAAN(ICANN-24)\cite{Su2024CrossModalAA}&\XSolidBrush&512& 0.560&0.665&0.528&0.624&0.730&{0.809}\\

    \textbf{RAMLN}&\XSolidBrush&512&\textcolor{red}{ 0.649}&\textcolor{red}{0.719}&\textcolor{red} {0.695}&\textcolor{red} {0.758}&\textcolor{red}{0.758}&\textcolor{blue}{0.815}\\
  \bottomrule
\end{tabular}}
\end{table*}

\subsection{Overall Objective}
The whole loss function $\mathcal{L}$ of our framework consists of two components: the embedding loss $\mathcal{L}_{{ra-qua}}$ and the classification loss $\mathcal{L}_{{cls }}$. So the loss function can be formulated as:
\begin{equation}\label{eq2_1}
\mathcal{L}=\mathcal{L}_{ra-qua} + \mathcal{L}_{cls}.
\end{equation}

Our method can be briefly summarized as Algorithm \ref{alg:algorithm}. It contains classification training and margin obtained through meta-optimization for metric learning.
\section{Experiments}
\subsection{Experimental Settings}

\textbf{Dataset Setting.} Sketchy \cite{sangkloy2016sketchy} is a fine-grained dataset and consists of 75,481 sketches and 12,500 photos from 125 categories. It has many instance-level matches. To achieve data balance, \cite{liu2017deep} expanded the photo gallery by collecting an extra 60,502 images. Then \cite{yelamarthi2018zero}  introduced a new protocol using 21 carefully selected categories, not present in ImageNet, to serve as unseen classes for testing purposes. Hence it is called Sketchy-NO, in which 21 unseen classes are used as unseen for testing and other 104 classes for training.

TU-Berlin \cite{eitz2012humans} contains 20,000 sketches over 250 object categories. And an extra 204,070 photos collected by \cite{liu2017deep} are included in the extended version. Unlike Sketchy, it is a category-level dataset. However, the number of photos is one-tenth that of sketches, thus it is imbalanced. According to many works in ZS-SBIR, the difficulty is higher than Sketchy. Following the partitioning protocol introduced in \cite{shen2018zero}, We choose randomly 30 classes as the unseen for testing, and the other 220 classes are used for training. We follow the conventional \emph{PK}sampling strategy \cite{hermans2017defense} to form batches by randomly sampling \emph{P} classes.

\noindent\textbf{Implementation details.} 
For comparison, we adopt two different pre-trained models as the backbone: CSE-ResNet50 and CLIP pre-trained ViT-B/32 image encoder. We find experimentally that large learning rates or frozen backbone achieve general results. The former may result from catastrophic forgetting. To ensure that the backbone network largely retains its original weights while adapting to the task of extracting features from the sketch domain, we assign a small learning rate to the backbone. Our model is trained with Adam \cite{kingma2014adam} optimizer with the learning rate of $2\times10_{}^{-5} $ for CLIP backbone network and the learning rate of CSE-ResNet is $1\times10_{}^{-3} $. The weight decays is $5\times10_{}^{-4} $. The learning rate of backbone is $7\times10_{}^{-3} $ of the main learning rate. The code is implemented with PyTorch \cite{paszke2017automatic} library and the experiments are conducted on NVIDIA GeForce RTX 3070 GPU. The batch size is set to 128 and the maximum number of training epochs is set to 20.

\noindent\textbf{Evaluation Protocol.}
We evaluate our model by adopting the evaluation protocol that most works adopt. For Tu-Berlin Extended dataset, we report the average of mean Average Precision (mAP@all) and Precision (Prec@100). And for Sketchy Extended dataset, we report the average of mean Average Precision (mAP@200l) and Precision (Prec@200).

\subsection{ Comparison with State-of-the-art Methods }
We evaluate our method against existing state-of-the-art approaches on both the Sketchy Extended dataset \cite{yelamarthi2018zero} and the TU-Berlin Extended dataset \cite{shen2018zero}. Two types of models are compared: those trained with semantic information and those without. Evidently, incorporating semantic domain knowledge enhances the sharpness of image feature extraction. Table \ref{tab-marks} presents the comparison with state-of-the-art methods, noting the differences in backbones. Earlier studies predominantly used ResNet or its variants, while recent works employ a larger ViT backbone with more parameters. As indicated in Table \ref{tab-marks}, our method outperforms all ViT-based models that exclude semantic information. Specifically, our approach achieves an 8.0\% and 17\% improvement in mAP over ZSE on the two datasets, as well as a 9.2\% and 8.0\% improvement over IVT. Our method also demonstrates strong performance with the ResNet backbone, achieving results comparable to state-of-the-art methods and demonstrating adaptability across backbones.  

Many works \cite{deng2020progressive, wang2021transferable, wang2021domain} go beyond learning the semantic data of the dataset itself. Some models also learn from additional linguistic datasets like WordNet. These semantic works are effective but costly. Our meticulously selected relation-aware quadruplets assist the image encoder in adapting to abstract sketches and aligning the two domains of natural photos and sketches with multiple constraints. We believe that our approach is simple and effective. All these results show that our method can effectively alleviate the modality gap between sketches and photos while reducing the large intra-class diversity in both photo and sketch domains.


%
%

\subsection{Ablation Experiment}
We conducted experiments with other components on TU-Berlin dataset. There are classification loss $\mathcal{L}_{{cls}}$, quadruplet loss $\mathcal{L}_{ {ra-qua}}$, and obtained margin $\mathcal{R}(x)$. As shown in Table \ref{tab-abl2}, our methods all contribute to the model. The combination of $\mathcal{L}_{{cls}}$ and $\mathcal{L}_{ {ra-qua}}$ has a greater improvement. This is due to $\mathcal{L}_{{cls}}$ helping $\mathcal{L}_{ {ra-qua}}$ to avoid local optimality. Only $\mathcal{L}_{ {ra-qua}}$ reduces the overall distance and make the feature distribution compact during training. $\mathcal{L}_{ {ra-qua}}$  suppresses this effect so that the model achieves better performance. Therefore, even if retrieval is based only on distance, classification training still makes sense. Margin $\mathcal{R}(x)$ obtained by meta-learning based on memory is also effective. It combines the advantages of long-term memory and short-term memory to match the current features with the vectors in the external memory matrix and makes reasonable predictions of margin values.
\begin{table}[!h]
\centering
\caption{Ablation studies for the proposed method on the TU-Berlin Extended dataset. The fixed margin is set to 0.3 when not using margin $\mathcal{R}(x)$ . }
\label{tab-abl2}
\setlength{\tabcolsep}{1mm}
\begin{tabular}{*{7}{c}}
  \toprule
  \multirow{2}*{$\mathcal{L}_{{cls}}$} & \multirow{2}*{$\mathcal{L}_{ {ra-qua}}$} & \multirow{2}*{ margin} &\multicolumn{2}{c}{TU-Berlin Ext.}\\
  \cmidrule(lr){4-5}    
  &  &   &mAP@all&Prec@100\\
  \midrule
   $\checkmark$& & & 0.449 &0.561\\
     & $\checkmark$& &0.481&0.571\\
    $\checkmark$&$\checkmark$ && 0.501&0.601\\
   &$\checkmark$&$\checkmark $&0.490&0.585\\
    $\checkmark$&$\checkmark $ &$\checkmark $&\textbf{0.509}&\textbf{0.615}\\
  \bottomrule
\end{tabular}
\end{table}

\subsection{Qualitative Analysis}
\begin{figure}[!h]
\centering
\includegraphics[width=0.49\textwidth]{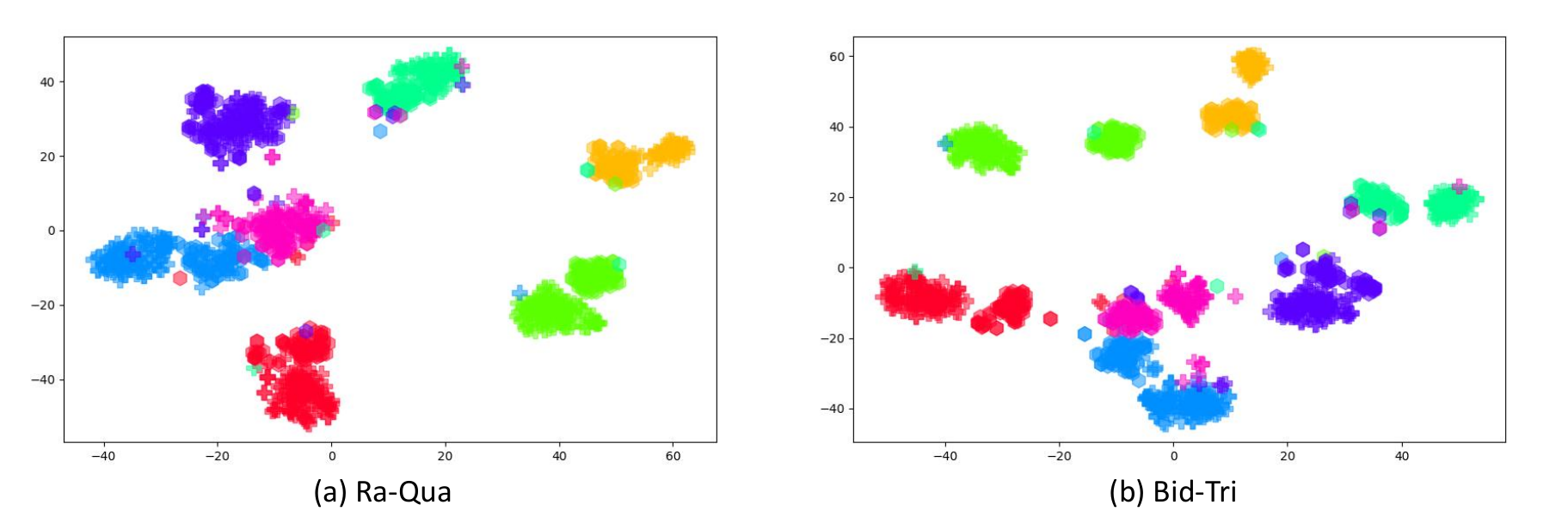}
\caption{T-SNE visualization of sketch and photo embeddings on Tuberlin Extended dataset. We randomly samples from 7
 test categories for visualization. Different colors refer to different categories. Crosses denote photos and hexagons denote sketches. (a) is our purposed method and (b) is \textbf{Bid-Tri}. It shows that our method has better clustering effect. }
\label{tsne}
\end{figure}

\textbf{Visualization of Feature Embedding.}
Figure. \ref{tsne} visualizes the distributions of 7 classes of Tuberlin by t-SNE which include seen and unseen classes. We compare our method with \textbf{Bid-Tri} as discussed in Section 2.1 of the Supplementary Materials. The optimization objective of  \textbf{Bid-Tri} is naive compared to our relation-aware quadruplet. This result shows that both methods have excellent ability to compress intra-class and separate inter-class features. In addition, for features from different modalities, the relation-aware quadruplet performs better. There is almost no gap between the groups from the sketch and the photo domain. It indicates that our method is also excellent in the distance control of both modalities. Features are well clustered together regardless of modalities. Also, all classes are separated by a certain distance. In addition, there are some images that are clustered into the wrong group. How to solve these difficult cases will become a research direction.

\begin{figure}[!h]
\centering
\hspace{-0.15cm}
\includegraphics[width=0.48\textwidth]{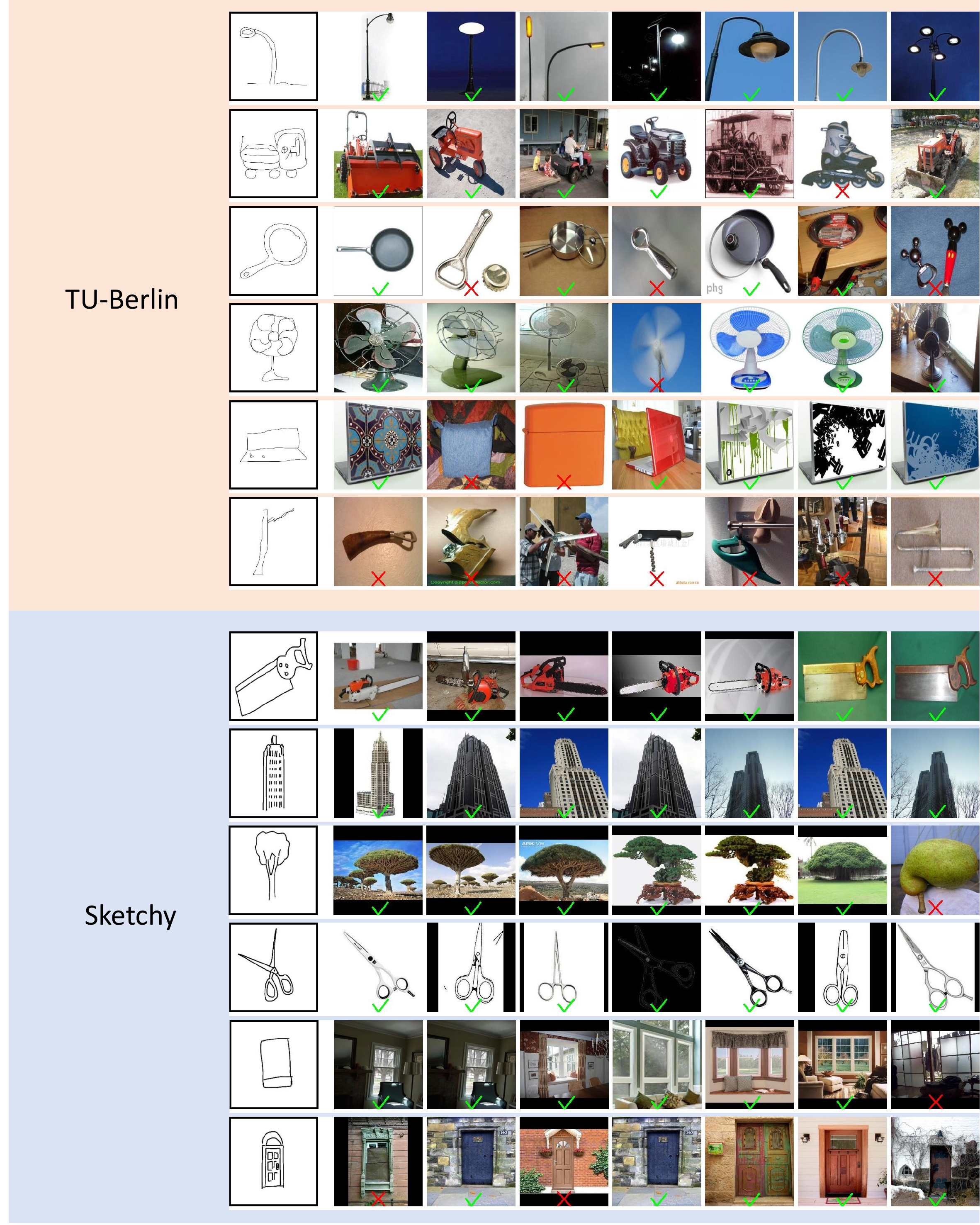}
\caption{Top 7 image retrieval examples of TU-Berlin and Sketchy. All of the examples come from unseen class. We use ticks and crosses to indicate right and wrong retrievals.
 }
\label{toplizi}
\end{figure}

\noindent\textbf{Visualization of Retrievals.}
We verify the ability of the proposed model on TU-Berlin and Sketchy. As shown in Figure. \ref{toplizi}, our model is successful for most samples, but it still fails to retrieve some candidates. We start by analyzing the data itself. It is easy to see that the two datasets have different characteristics. Sketchy has more details and is more similar to real objects. These may be due to differences in the requirements of the participants and their own painting levels. It is also why Sketchy is a fine-grained dataset, but many works achieve higher result in Sketchy than that in TU-Berlin for the same evaluation protocol.

\section{Conclusion}
We introduce a metric learning framework leveraging a relation-aware quadruplet loss to capture both inter and intra-modal relationships in the ZS-SBIR task. It takes into account the features of ZS-SBIR and uses two modal contrastive learning to mitigate the effect of modal gaps. Additionally, it prevents feature overlap between samples from different classes within the same modality, resulting in superior performance compared to other triplet-like loss designs. The adaptive margin derived from our meta-learning strategy removes the need for manual margin tuning. By adjusting automatically, it mitigates issues caused by suboptimal configurations. This adaptive margin also accounts for domain variations across categories and modalities. As a result, it significantly improves the generalization capacity of metric learning. Experimental results on the TU-Berlin Extended and Sketchy Extended datasets confirm the effectiveness of our proposed method for cross-modal retrieval. The method performs well with object sketches of varying granularity. It also demonstrates robustness against variations in object shape and sample diversity.
{
    \small
    \bibliographystyle{ieeenat_fullname}
    \bibliography{main}
}

\end{CJK}
\end{document}